# Bilingual Terminology Extraction Using Multi-level Termhood

Chengzhi ZHANG [a], Dan WU [b]

a. Department of Information Management, Nanjing University of Science and Technology, 210093, Nanjing, China
b. School of Information Management, Wuhan University, 430072, Wuhan, China
zhangchz@istic.ac.cn, woodan@whu.edu.cn

## ABSTRACT

**Purpose**: Terminology is the set of technical words or expressions used in specific contexts, which denotes the core concept in a formal discipline and is usually applied in the fields of machine translation, information retrieval, information extraction and text categorization, etc. Bilingual terminology extraction plays an important role in the application of bilingual dictionary compilation, bilingual Ontology construction, machine translation and cross-language information retrieval etc. This paper addresses the issues of monolingual terminology extraction and bilingual term alignment based on multi-level termhood.

**Design/methodology/approach**: A method based on multi-level termhood is proposed. The new method computes the termhood of the terminology candidate as well as the sentence that includes the terminology by the comparison of the corpus. Since terminologies and general words usually have differently distribution in the corpus, termhood can also be used to constrain and enhance the performance of term alignment when aligning bilingual terms on the parallel corpus. In this paper, bilingual term alignment based on termhood constraints is presented.

**Findings**: Experiment results show multi-level termhood can get better performance than existing method for terminology extraction. If termhood is used as constrain factor, the performance of bilingual term alignment can be improved.

**Originality/value**: The termhood of the candidate terminology and the sentence that includes the terminology is used to terminology extraction, which is called multi-level termhood. Multi-level termhood is computed by the comparison of the corpus. The experiment results show that the multi-level termhood can get better performance than standard method. Bilingual term alignment method based on termhood constraint is put forward and termhood is used in the task of bilingual terminology extraction. Experiment results show that termhood constraints can improve the performance of terminology alignment to some extent.

# 1. INTRODUCTION

Terminology is the product of scientific and technological development. It is the set of technical words or expressions used in specific contexts, which denotes the core concept in a formal discipline. It can be applied in the research area of natural language processing (NLP), information retrieval, machine translation and data mining. There are two characters of terminology. On the one hand, terminology denotes the core concept in a discipline so the usage and the users are limited. General words are more acceptable and recognizable in different areas than terminology. On the other hand, compared to general words, terminology always has a single meaning in specific domain. These two characters of terminology can be computed by termhood. The higher the termhood is, the higher capability of distinguishing different domains the terminology has.

In the previous works about terminology extraction, only the termhood of a terminology candidate is considered, but not other aspects. In this paper, the termhood of the candidate terminology and the sentence that includes the terminology is used to terminology extraction, which is called multi-level termhood. Multi-level termhood is computed by the comparison of the corpus. The experiment results show that the multi-level termhood can get better performance than standard method.

Bilingual terminology extraction is composed of two steps: 1. extracting monolingual terms from monolingual document or sentence; 2. aligning the bilingual terms. Different from traditional word alignment, extracting and aligning bilingual terms from parallel sentences aims at extracting and aligning not all words but candidate terms in sentences. Therefore traditional word alignment methods can not directly applied to bilingual term alignment task. For those differently distributed terms and general words on corpus, result of traditional word alignment methods can be optimized or constrained with termhood during the process of term alignment to enhance the performance of term alignment. Bilingual term alignment method based on termhood constraint is put forward and termhood is used in the task of bilingual terminology extraction. Experiment results show that termhood constraints can improve the performance of terminology alignment to some extent.

The rest of this paper is organized as follows. The next section reviews some related work on bilingual terminology extraction. In section 3, a detailed description of terminology extraction based on multi-level termhood is presented. In section 4, the method of bilingual terminology alignment based on termhood constraints is described. The paper is concluded with a summary and directions for future work.

## 2. RELATED WORK

### 2.1. Terminology Extraction

With the enrichment of language resources and the development of NLP, many terminology extraction systems have been developed (Kit & Liu, 2008). The most commonly used terminology extraction methods include linguistic, statistics and hybrid approach.

**(1) Linguistic Approach**

Linguistic features are used to restrain the candidate terminology, that is, terminology is filtered by linguistic features (Ido and Ward, 1994). Linguistic methods exploit part-of-speech Tagging and shallow parsing to filter the terminology. (Bourigault, 1992) used shallow parsing to extract noun phrase that is a terminology. (Ido and Ward, 1994) limited the candidate terminology to a string that represents the pattern of noun sequences. (Justeson and Katz, 1995) used the prefix of terminology and selected strings whose prefix is noun to be candidate terms. Good results can be achieved in small corpora using linguistic methods, yet the recall rate is low for the shortage of patterns and the adaptability of fields and languages.

**(2) Statistical Approach**

Statistical approaches are based on the statistical information, such as the frequency of terms appearing in the corpora, to extract terms, including TF*IDF (Maedche & Staab, 2000), KF*IDF (Xu et al, 2002), C-value/NC-value (Frantzi et al, 2000) and so on.

Termhood extraction relates to two basic statistical variables, that is, Unithood and Termhood of terminology. Statistical method is usually used to compute these two variables. For unithood, the approaches includes MI (Church & Hanks, 1990), LogL (Dunning, 1993) and left/right entropy (Patry & Langlais, 2005). When computing termhood, methods such as TF*IDF (Maedche & Staab, 2000), DR-D (Velardi, 2001)、

C-value/NC value (Frantzi et al, 2000), inter-domain entropy (IDE) (Chang, 2005) and Domain Component Feature Set (DCFS) (Zhang et al, 2003) are employed.

**(3) Hybrid Approach**

Linguistics and statistics have their own advantages and disadvantages, and they are usually integrated to extract terminology. There are two ways to combine them. One way is to extract candidate terms with linguistics methods and then with the statistics methods. The other way is to obtain candidate terms using statistics methods first and then use linguistic methods to abandon those terms inconsistent with linguistic patterns.

(Daille, 1996) used the linguistic methods to get candidate terms and set them as the input of statistical models. Then statistical methods such as MI and LogL are used to get final terms. (Maynard & Ananiadou, 2000) did some research about the extraction of multi-word term using thesaurus and semantic Web to get the semantic and category information, and then integrated it with the statistical and syntactic information in the corpora.

Different terminology extraction toolkits can also be integrated to extract terminology besides the integration of linguistics and statistics methods (Kit & Liu, 2008). (Vivaldi & Rodríguez, 2000) integrated different term extraction tools by simple voting and the result is better than single term extraction tools. (Vivaldi & Màrquez, 2001) improved the above voting approach, got the best integration strategy by Boosting algorithm and enhanced the performance of terminology extraction based on hybrid approach.

## 2.2. Optimization of Bilingual Terminology Extraction

(Wu & Wang, 2004; Wu et al, 2005) did some research about optimization of domain terms alignment with large general parallel corpus. (Wu & Wang, 2004) trained models for word alignment using domain specific and general corpus separately and employed the two models to improve performance of word alignment. (Wu et al, 2005) further optimized their method further, changed self-adaptive methods into statistical models and improved the performance of word alignment.

It's important to note that domain term alignment optimization methods adopted by (Wu & Wang, 2004; Wu et al, 2005) rely on the availability of large domain and general parallel corpus. Their method has limitation when a large parallel corpus is not available. In this paper, termhood of terms in parallel sentences will be used instead of large general parallel corpus. General corpus of different languages is employed to compute termhood and results of term alignment can be optimized with termhood. So methods presented in this paper are expansive. (Wu & Wang, 2005; Wu et al, 2006) used integrated learning methods (including Bagging, Boosting and semi-supervised Boosting) to improve the result of word alignment.

Relevant research includes the study on confidence measurement of word alignment by (Huang, 2009). He introduced sentence alignment confidence measure and alignment links confidence measure to improve performance of word alignment by selecting aligned sentence and linked word with high confidence.

Some illegal sentence pairs (including some incorrect alignment result or partial alignment result) can be found by measuring termhood of sentences and evaluating alignment quality of parallel sentences pairs. Constraining results of term extraction with termhood can improve the performance of term extraction and alignment.

When studying evaluation method of word alignment, (Huang et al, 2009) pointed out that lack of links is less important than wrong links when aligning words. So it's valuable to find wrong results among results of word alignment.

## 3. TERMINOLOGY EXTRACTION BASED ON MULTI-LEVEL TERMHOOD

### 3.1. Features used in Terminology Extraction

In this section, CRF model is used to extract terminology from documents. CRF is a model of probability graph proposed by (Lafferty, 2001) which is widely used in word segmentation, part-of-speech tagging, chunking recognition, named entity recognition and so on. The features used in CRF model are shown in Table1.

In this paper, Segtag (a tool with function of Chinese segmentation and POS tagging, it downloaded from http://www.nlp.org.cn) will be used to segment Chinese documents and tag POS with a general word segmentation lexicon. Since the linguistics and statistics methods are integrated in the tool, words, POS and termhood will be used as features of CRF and models will be trained with training data.

Note that the above-mentioned POS refers to results of POS tagging of each segmented unit. These units may be part of a terminology or even not be a terminology.

CRF++ (http://crfpp.sourceforge.net) will be used in this paper to train and test the terminology extraction models. Statistical feature employed in this model, that is termhood of terminology and the sentence including this term will be illustrated.

### 3.2. Multi-level Termhood for Terminology Extraction

The linguistics and statistics features of candidate terminology will be integrated in this paper. Termhood of candidate terminology and sentences containing these terms will be considered synthetically.

**(1) Termhood of Candidate Terminology**

In this paper, CRF model will be used to extract terms on manual-tagged corpus with 10-fold cross-validation. Results show that methods based on frequency difference is better than those based on frequency rank difference. Both of them are dependent on the size of domain and general corpus. Therefore, next step is to find corpus comparison method based on frequency difference or rank difference.

Different from (Kit & Liu 2008), terminology extraction here is not limited to mono-word term extraction in this paper. When using CRF to extract terms, after text segmentation, part-of-speech tagging and computation of termhood, all of these features will be as input of CRF, including extraction of mono-word and multi-word terms.

In this paper, termhood is used to extract terminology. According to the research result of (Liu & Kit, 2008) about termhood computing of mono-word term and our preliminary experiment result, we compute termhood based on the frequency difference and rank difference between domain and general corpus. The brief description of the method is as follows.

Assumption: candidate term is w, corpus is x, total number of words in corpus x (size of lexicon generated by x) is $V_x$, $f_x(w)$ is the relative frequency of candidate term w occurred in x, $c_x(w)$ is the absolute frequency of candidate term w occurred in x, and $f_x(w)$ can be calculated by formula (1) (Liu & Kit, 2009).

$$f_x(w) = c_x(w) \Big/ \sum_{w' \in V_x} c_x(w') \qquad (1)$$

Frequency difference of candidate term w occurred in domain corpus $x_d$ and general corpus $x_b$ is computed by formula (2) (Liu & Kit, 2009).

$$\Delta f(w) = f_d(w) - f_b(w) \tag{2}$$

Rank difference of candidate term w occurred in domain corpus $x_d$ and general corpus $x_b$ is computed by formula (3) (Liu & Kit, 2009).

$$\Delta r(w) = r_d(w) - r_b(w) \tag{3}$$

And, $r_d(w)$、$r_b(w)$ are the rankings of frequency of term w occurred in domain corpus $x_d$ and general corpus $x_b$ (Liu & Kit, 2009) (ranking can be reversed, that is, the higher the ranking is, the larger the number is (Kit & Liu, 2008)). In this paper, tagged data of People's Daily (http://www.people.com.cn/) from January to June 1998 are used as general corpus to extract terms.

On the basis of rank difference, number-intensified is done by formula (4) (Liu & Kit, 2009) to get new rank difference after the enhancement.

$$\Delta r_c(w) = c_d(w) \cdot (r_d(w) - r_b(w)) \tag{4}$$

**(2) Termhood of Sentences Containing Candidate Terminology**

Termhood of sentences refers to mean value of termhood of all words which are in the same sentences as the candidate terms. In this paper, corpus such as journal article title, journal article abstract, patent title, patent abstract, news title, news article, MARC (Machine-Readable Catalogue) title and MARC summary is used as statistical sample to compute termhood and average sentence termhood of diverse corpus.

As shown in table2, from the perspective of title, ranking of termhood of different corpus is: patent corpus> journal article corpus > MARC corpus > news corpus.

From the perspective of full text or abstract, ranking of termhood of different corpus is: MARC corpus > journal article corpus > patent corpus > news corpus.

Previous research results about automatic keyword extraction show that title contributes more than abstract and full texts for keyword extraction (Hou et al, 2005). But term extraction and keyword extraction are two different tasks. Keyword extraction aims at extracting 6 to ten words that mostly represent the content of the document. However, term extraction extracts terms in certain documents or collections and the number of extracted terms depends on the document itself without any restriction. Compared to abstract and full text, title is the summary of full text, and contains less terms in order to increase readability. Abstract is the summary of key points of the document, so abstracts and full texts contain more terms. Results of sampling statistics also demonstrate this point. What's more, the difference between title and abstract is more significant in MRAC than in dissertation. In future, more data will be used as statistical samples to analyze termhood differences of titles, abstracts and full texts.

As shown in the statistical result, the sentence termhood of title and full text in news is lower than that of specific domain. In real texts, professional literature usually contains more terms while news has more general words. So when extracting terms, termhood of sentence can be integrated into term extraction model as part information of candidate term.

Termhood of term itself and that of sentences or articles containing this term can complete mutual learning, that is, the higher termhood of sentences or articles is, the more likely that they contain terms and the higher termhood of a term is, the higher termhood of sentences or article containing this term is. The two kinds of information can be mutually learned through interaction to enhance the quality of term extraction. What's more, terms are more likely to appear in sentence whose termhood is high. Termhood of all sentences which contain this term in corpus can be used as a global feature to complete data training.

It's important to note that recently mutually reinforcing relationship between domain sentences and domain terms are studied by (Yang et al, 2010) to extract terms. They implemented this mutual reinforcing relationship by Link Analysis. Different from their work, in this paper, multi-level termhood such as term termhood and sentence termhood is integrated in the term extraction model trained by CRF model.

### 3.3. Experiment and Result Analysis

**(1) Training Data and Evaluation Method**

For lack of tagged English corpus, we only did experiment on Chinese text corpus to extract term with CRF model and several factors that influence the performance of term extraction is tested and analyzed.

Tagged Chinese corpus used in this paper is mainly article about computer and it is manually tagged with BIO (Begin, In, Out) tag, containing 1334 sentences, 15172 words and punctuations. There are 1910 manual-tagged terms (including repeated terms).

The common used evaluation merit for term extraction is precision, recall and $F_1$ value. Assuming that there are n words in test set, extraction result can be represent as shown in table 3. This experiment splits manually-tagged results into two groups, that is terms manually tagged (word or phrase responsive to tag sequence "B-I-…") and non-terms tagged manually (word or phrase before label "O").

- **Precision**

$$P = \frac{a}{a+b} \tag{5}$$

Precision P is the precision ration of term extraction. Precision indicates the ability of term extraction to get correct terms. The higher precision is, the less likely a term is a non-term.

- **Recall**

$$R = \frac{a}{a+c} \tag{6}$$

Recall R is the ratio of tagged term and it indicates the ability of the term extraction system to find terms. The higher recall is, the less terms are untagged.

- $F_1$ value

$$F_1(P,R) = \frac{2PR}{P+R} \tag{7}$$

$F_1$ measurement is presented by van Rijsbergen which is the harmonic-mean of precision and recall (van Rijsbergen, 1979).

10-fold cross-validated method is used in the experiment and P, R and $F_1$ are employed for evaluation.

**(2) Experiment Results and Analysis**
- **Result and Analysis of Multi-level Termhood**

Table 4 is the result of term extraction experiment that integrates multi-level termhood features.

Table 4 shows that precision is improved when termhood of words in term is considered.

When the simplest corpus comparison method, which uses frequency to compare domain and general corpus, is used, precision is 4% higher, and recall is 1% higher. When ranking value based corpus comparison method is used, the precision is 3% higher while recall drops greatly. When the way of difference is considered, corpus comparison method based on frequency and ranking is improved.

The performance of term extraction can also be improved but not significantly when ranking value is enhanced by number.

Table 4 shows that on the basis of frequency difference or rank value difference, when termhood of sentence (ΔFreq_Sen or ΔRank_Sen) which contains current candidate term is employed, $F_1$ has no obvious change, yet recall is 1% higher. This illustrates that more terms can be found when termhood of sentences are considered as well as termhood of terms. And, among all termhood methods, the one which employs frequency difference of candidate terms and cumulative frequency difference of sentences which contains candidate terms get the highest $F_1$ value and best performance.

Precision of term extraction improves slightly when frequency or ranking value of the candidate term on domain and general corpus was integrated with their difference.

Among all termhood methods, the one that integrates frequency difference, ranking difference of candidate term in domain and general corpus, and cumulative frequency difference and ranking difference of the sentences that contains the candidate terms get the highest recall.

Experiment result shows that termhood can enhance the performance of term extraction. Multiple differences can improve the precision of term extraction. Recall can be improved when termhood of sentences is employed.

- **Result and Analysis of Different Features and Combination of Features**

Table 5 shows the result getting from different feature combination. It illustrates that the method employing word as the only feature performed the worst considering precision and recall. When POS was integrated, the results were improved. The performance was enhanced significantly when the frequency of word appearing in the domain and general corpus was considered.

Among all feature combinations, the methods employed POS, frequency, frequency difference and ranking difference of candidate terms on domain corpus and general corpus has the highest precision. The methods employed frequency difference and ranking difference of candidate terms on domain corpus and general corpus, and cumulative frequency difference and ranking frequency of sentences which contains candidate terms achieved the highest recall. When frequency difference of candidate terms and cumulative frequency difference of sentences which contains candidate terms was used in extraction model, the corresponding $F_1$ value is the highest.

## 4. BILINGUAL TERMINOLOGY ALIGNMENT BASED ON TERMHOOD CONSTRAINTS

### 4.1. Using Termhood to Optimize the Bilingual Terminology Alignment

The key idea of extracting bilingual terms from bilingual sentences-aligned corpus is that relevance is computed on the basis of co-occurrence of bilingual terms in a bilingual sentence-aligned corpus. POS and word frequency are often used in the process of bilingual word alignment to filter bilingual candidate terms. If the term in one language has high termhood, the corresponding term in other languages should also have high termhood.

According to this assumption, termhood of Chinese and English candidate terms are used as constraints to study the alignment of Chinese-English terms. When computing relevance of bilingual terms, termhood and termhood ratio of Chinese-English terms are added as weights to compute the results of relevance.

Assume Chinese word c, English word e, and their termhood, i.e. *Termhood(c)* and *Termhood(e)*, after termhood ratio is integrated, their relevance which denoted as *Association(c, e)*, turns to:

$$Association(c,e)' = \frac{Association(c,e)}{Max\{\frac{Termhood(e)}{Termhood(c)}, \frac{Termhood(c)}{Termhood(e)}\}} * Termhood(c) * Termhood(e)$$

(8)

Given threshold $\theta$, word pairs which satisfy $Association(c,e)' \geq \theta$ or $Association(c,e)'$ are among the first k one is set to be the candidate term pairs.

The procedure of bilingual term alignment based on termhood constraints is as follows:

After extracting bilingual terms from a domain parallel corpus, termhood is computed with Chinese-English bilingual general corpus and the relevance of Chinese-English terms is computed using termhood and termhood ratio as weights, and the alignment results with the first N relevance are the candidate bilingual term pairs. Bilingual termhood ratio can be used to constrain further on the basis of bilingual term alignment to filter candidate term pairs with termhood difference.

### 4.2. Result and analysis of term extraction

**(1) Test Corpus and Evaluation Method**

The general corpus used in this experiment was the same one used in the project that tagged data from People's Daily between Jan 1998 and Jun 1998 and English news corpus of NTCIR (http://research.nii.ac.jp/ntcir/) in 1998-2001. Domain corpus contains disserations in the infomation technology domain. More detail about corpus is shown in table 6.

The evaluation measure used in this paper is precision at first N words, denoted as *P@N*. As shown in formula (9), *P@N* examines the ratio of correct alignments in the first N word alignment results

$$P@N = \frac{Number\ of\ correct\ alignment\ of\ the\ first\ N\ alignment\ result}{N} \quad (9)$$

Two types of checks were manually performed on the first N alignment results. First, four groups of bilingual term extraction experiments were conducted using 1000, 2000, 5000, and 10000 sentence pairs separately extracting from IT parallel corpus. In each group, extraction models used included N-Gram and CRF model (both of them are employed in Chinese term extraction and only N-Gram is used in English term extraction). Statistical relevance of bilingual terms was computed using Log likehood ratio (Dunning, 1993) and term-weighted Log likehood ratio.

Relevance of bilingual candidate terms was computed by LogL likehood ratio. Given Chinese word c, English word e and relevance *LogL(c, e)*, term-weighted LogL is

$$LogL(c,e)' = \frac{LogL(c,e)}{Max\{\frac{Termhood(e)}{Termhood(c)}, \frac{Termhood(c)}{Termhood(e)}\}} * Termhood(c) * Termhood(e) \quad (10)$$

Second, the results of Chinese-English term alignment and Chinese term extraction of terms with the first 500 relevance were manually annotated by four volunteers. The annotation was performed on the results of bilingual term extraction in the four corpora with different sizes. Based on the manual annotation, if the Chinese word is a term, the alignment of Chinese-English words was added and normalized, and precision *P@N* of Chinese-English term extraction was computed.

**(2) Analysis of Experiment Result**

Experiments in this section were carried out according to each scheme to obtain bilingualterm extraction and alignment results. Results of differnet scheme were manually evaluated according to the above-mentioned methods, and were compared and analyzed from two sides.

Table 7 shows the precision of word alignment on parallel corpus with different sizes. Without termhood constraints, precision of word alignment using N-Gram model is actually better. However, termhood constraints have opposite effect on CRF model. It shows that the precision of alignment can be enhance by weighted termhood. By analyzing extraction results, we found that N-Gram model extracted more non-term words.

## 5. CONCLUSION AND FUTURE WORKS

This paper discusses an innovative method for extracting bilingual terminology based on multi-level termhood. The multi-level termhood includes termhood of the terminology candidates and the sentence, and it is computed by the comparison of the corpus. This paper also put forward bilingual word alignment based on termhood constraint, and the termhood is used in the task of bilingual term extraction and alignment. Experiment results show that termhood constraints can enhance performance of term alignment.

The future work include: finding a method to compare corpora that are not limited by domain and background in computing termhood of candidate terminology; adding more features such as mutual information among words, chunking and semantics information to improve the precision of terminology extraction; using the Web to obtain background corpora to compute termhood and improve the recall of certain domain; employing integrated learning methods in the process to enhance the performance of terminology extraction system.

The future work that will be done about bilingual terminology alignment based on termhood constraint includes: optimizing results of bilingual term alignment with multi-level termhood; using more statistical relevance methods in computing multiple relevance of bilingual terms to have an integrated method for improving the precision of bilingual term alignment.

Table1. Features of candidate terminology

| No. | Type of feature | Tag of feature | Meaning of feature |
|---|---|---|---|
| 1 | Basic feature | *Word* | Word itself |
| 2 | | *Len* | Length of Word |
| 3 | | *POS* | Part-of-speech of Word |
| 4 | | *Count* | Number of words of the sentence including Word |
| 5 | | *Freq_D* | Frequency of Word in domain corpora |
| 6 | | *Freq_B* | Frequency of Word in background corpora |
| 7 | | *Rank_D* | Rank value of frequency of Word in domain corpora |
| 8 | | *Rank_B* | Rank value of frequency of Word in background corpora |
| 9 | | *Freq_Sen_D* | Sum of frequency of words which are in the same sentence with Word in domain corpora |
| 10 | | *Freq_Sen_B* | Sum of frequency of words which are in the same sentence with Word in background corpora |
| 11 | | *Rank_Sen_D* | Sum of rank value of words which are in the same sentence with Word in domain corpora |
| 12 | | *Rank_Sen_B* | Sum of frequency of words which are in the same sentence with Word in background corpora |
| 13 | Combined feature | *Δ Freq* | Freq_D- Freq_B |
| 14 | | *ΔRank* | Rank_D- Rank_B |
| 15 | | *ΔFreq_Sen* | Sum of frequency difference of words which are in the same sentence with Word between domain corpora and background corpora |
| 16 | | *ΔRank_Sen* | Sum of rank value difference of words which are in the same sentence with Word between domain corpora and background corpora |

Table2. Comparison of sentence termhood of different domain corpus

| Document Type | Type of domain | Number of sentences | Average Frequency of domain corpus | Average Frequency of general corpus | Mean Termhood (↓) |
|---|---|---|---|---|---|
| Summary of MARC | economics | 2000 | 237.7135897 | 52613.96954 | 0.004518070 |
| Abstract of journal article | Social science | 1050 | 95.75946343 | 33699.52687 | 0.002841567 |
| Abstract of patent | electric elements | 1070 | 87.4899300 | 30917.43000 | 0.002829793 |
| Abstract of dissertation | Information technology | 2000 | 123.4161849 | 45835.41233 | 0.002692595 |
| Title of dissertation | Information technology | 2000 | 135.1281599 | 50860.22811 | 0.002656853 |
| Title of patent | New energy automobiles | 1000 | 109.8355309 | 41619.10867 | 0.002639065 |
| Title of journal article | Social science | 1000 | 100.7394858 | 41837.64822 | 0.002407867 |
| Full text of news | news | 2000 | 104.0241397 | 47291.66036 | 0.002199630 |
| Title of MARC | economics | 2000 | 81.75716858 | 41466.33512 | 0.001971652 |
| Title of news | news | 1000 | 61.81313494 | 34242.90108 | 0.001805137 |

Table3. Contingency table for evaluation of tagged result

|  | Term tagged manually | Non-term tagged manually |
|---|---|---|
| Term tagged by extraction system | a | b |
| Non-term tagged by extraction system | c | d |

Table4. Effect of different termhood computing methods for terminology extraction

| Measurement of Termhood | P | R | $F_1$ |
|---|---|---|---|
| No Measurement of Termhood | 0.80098 | 0.7422 | 0.76851 |
| *Freq_D, Freq_B* | 0.83870 | 0.75184 | 0.79126 |
| *Rank_D, Rank_B* | 0.82822 | 0.63861 | 0.71749 |
| *Freq_D, Freq_B, Rank_D, Rank_B* | 0.84265 | 0.74458 | 0.78806 |
| *Δ Freq* | 0.84098 | 0.75443 | 0.79385 |
| *ΔRank* | 0.84036 | 0.75225 | 0.79221 |
| *Number-intensified ΔRank* | 0.84144 | 0.75648 | 0.79508 |
| *Δ Freq, ΔFreq_Sen* | 0.83275 | 0.76563 | 0.79638 |
| *ΔRank, ΔRank_Sen* | 0.82371 | 0.76247 | 0.7904 |
| *Freq_D, Freq_B, Δ Freq, ΔRank* | 0.84326 | 0.74787 | 0.79118 |
| *Rank_D, Rank_B, Δ Freq, ΔRank* | 0.84137 | 0.74107 | 0.78642 |
| *Δ Freq, ΔRank, ΔFreq_Sen, ΔRank_Sen* | 0.83007 | 0.76755 | 0.79626 |

Table5. Effect of different features and combination of feature to terminology extraction

| Features | P | R | $F_1$ |
|---|---|---|---|
| Word | 0.80098 | 0.7422 | 0.76851 |
| Word, POS | 0.81183 | 0.63934 | 0.71136 |
| Word, Freq_D, Freq_B | 0.81183 | 0.63934 | 0.71136 |
| Word, Freq_D, Freq_B, POS | 0.83870 | 0.75184 | 0.79126 |
| Word, Δ Freq, ΔFreq_Sen, POS | 0.83275 | 0.76563 | 0.79638 |
| Word, Freq_D, Freq_B, Δ Freq, ΔRank, POS | 0.84326 | 0.74787 | 0.79118 |
| Word, Δ Freq, ΔRank, ΔFreq_Sen, ΔRank_Sen, POS | 0.83007 | 0.76755 | 0.79626 |

Table6. Test Corpus for bilingual terminology extractor and alignment

| Type of Corpus | | Description | Total documents | Total words | Total of different words |
|---|---|---|---|---|---|
| Chinese | Domain | disserations in IT | 20, 000 | 7, 795, 569 | 74, 308 |
| | Back-ground | People's Daily (Jan~Jun, 1998 | 18, 670 | 7, 286, 875 | 141, 989 |
| English | Domain | disserations in IT | 20, 000 | 6, 953, 675 | 169, 491 |
| | Back-ground | News corpus of NTCIR (1998-2001) | 75, 007 | 34, 016, 881 | 433, 496 |

Table7. Precision of bilingual term alignment is P@500

| Model<br>Number of sentence) | LogL Relevance | | LogL relevance with weighted termhood | |
|---|---|---|---|---|
| | N-Gram | CRF | N-Gram | CRF |
| 1, 000 | 0.5140 | 0.3750 | 0.4450 | 0.4702 |
| 2, 000 | 0.4840 | 0.3690 | 0.4460 | 0.4130 |
| 5, 000 | 0.5690 | 0.3500 | 0.4830 | 0.3739 |
| 10, 000 | 0.6840 | 0.4820 | 0.6610 | 0.5500 |